\documentclass[preprint]{elsarticle}
\makeatletter
\def\ps@pprintTitle{%
 \let\@oddhead\@empty
 \let\@evenhead\@empty
 \def\@oddfoot{}%
 \let\@evenfoot\@oddfoot}
\makeatother
\journal{Journal of Computational Physics}

\usepackage{xcolor,color}
\usepackage{tikz}
\usepackage{units}
\usepackage{amsmath}
\usepackage{amssymb}
\usepackage{amsfonts}
\usepackage{courier}
\usepackage{bm}
\usepackage{hyperref}
\usepackage{cleveref} % use command \cref 

\usepackage{amsthm}
\theoremstyle{remark}
\newtheorem{remark}{Remark}

\usepackage{algorithm}
\usepackage{algpseudocode}
\usepackage{pifont}
\usepackage{natbib}
\usetikzlibrary{external}
\definecolor{textblue}{HTML}{1F4C84}

\begin{document}

\graphicspath{
               {Figures/}
              }
              
%%\maketitle              
         
\begin{frontmatter}



\title{Data-Driven Discovery of Coarse-Grained Equations} 

%%\author{Joseph Bakarji \and Daniel M. Tartakovsky}
\author[label1]{Joseph Bakarji}
\author[label2]{Daniel M. Tartakovsky}
\ead{tartakovsky@stanford.edu}

\address{Department of Energy Resources Engineering, Stanford University, 367 Panama Mall, Stanford, CA 94305, USA }

\begin{abstract}
Statistical (machine learning) tools for equation discovery require large amounts of data that are typically computer generated rather than experimentally observed. Multiscale modeling and stochastic simulations are two areas where learning on simulated data can lead to such discovery. In both, the data are generated with a reliable but impractical model, e.g., molecular dynamics simulations, while a model on the scale of interest is uncertain, requiring phenomenological constitutive relations and ad-hoc approximations. We replace the human discovery of such models, which typically involves spatial/stochastic averaging or coarse-graining, with a machine-learning strategy based on sparse regression that can be executed in two modes. The first, direct equation-learning, discovers a differential operator from the whole dictionary. The second, constrained equation-learning, discovers only those terms in the differential operator that need to be discovered, i.e., learns closure approximations. We illustrate our approach by learning a deterministic equation that governs the spatiotemporal evolution of the probability density function of a system state whose dynamics are described by a nonlinear partial differential equation with random inputs.  A series of examples demonstrates the accuracy, robustness, and limitations of our approach to equation discovery.
\end{abstract}

\begin{keyword}
machine learning \sep closure approximation \sep coarse-graining \sep stochastic

\end{keyword}
\end{frontmatter}


\section{Introduction}

Empiricism, or the use of data to discern fundamental laws of nature, lies at the heart of the scientific method. With the advent of ``machine learning'', this ancient facet of pursuit of knowledge takes the form of inference, from observational or simulated data, of either analytical relations between inputs and outputs~\citep{schmidt2009distilling} or governing equations for system states~\citep{geneva2020modeling, pang2019fpinns, raissi2017machine, raissi2019physics, sirignano2018dgm, wu2020data}.
%With the advent of ``machine learning'', this ancient facet of pursuit of knowledge takes the form of inference from observational or simulated data. A data-driven relationship between the inputs and the outputs takes the form of an analytical expression, e.g.,~\citep{schmidt2009distilling}, or a governing equation---ordinary or partial differential equations (ODEs or PDEs)---for a state variable~\citep{geneva2020modeling, pang2019fpinns, raissi2017machine, raissi2019physics, sirignano2018dgm, wu2020data}.
 The advantage of learning a governing equation, rather than an input-output map for a quantity of interest (QoI), is the possibility to ``generalize'' (extrapolate) over space and time and over different external inputs such as initial and boundary conditions. %Space-time generalization in physics is often ignored when applying machine learning methods to physical problems. 
In this sense, learning a differential equation is akin to learning an iterative algorithm that generates a solution, rather than learning the solution itself. Of direct relevance to the present study is the use of sparse regression on noisy data to estimate the constant coefficients in nonlinear ordinary~\citep{Brunton2016} and partial~\citep{Rudy2017, Schaeffer2017} differential equations (ODEs and PDEs, respectively). This strategy has been generalized to recover variable coefficients~\citep{Rudy2019} or nonlinear constitutive relations between several state variables~\citep{Tartakovsky-2020-physics}.
%It has also been shown that learning equations can be done in the context of deep learning by carefully choosing the activation functions as a basis \citepp{Martius2019}.

In physical sciences, observational data are seldom, if ever, sufficient to accomplish this goal; instead, the data must be generated by solving a governing equation. This strategy provides a partial explanation for why machine learning methods are yet to discover new physical laws: to generate the data one needs to know the underlying equation, which is subsequently learned from these data. Multiscale modeling and stochastic simulations are two areas where learning on simulated data can lead to real discovery. In multiscale simulations, one is reasonably sure of an appropriate model at one scale (e.g., basic laws of molecular dynamics) and aims to learn a model at another scale (e.g., a continuum-scale PDE) from the data generated at the first scale. Examples of machine learning techniques for upscaling, i.e., discovery of coarse-grained dynamics from fine-grained simulations, and downscaling, i.e., discovery of fine-grained dynamics from coarse-grained simulations, can be found in~\citep{boso2018information, felsberger2019physics, geneva2019quantifying, schoberl2017predictive}.

In stochastic simulations, one deals with governing equations that either contain uncertain (random) parameters or are driven by randomly fluctuating forcings that represent sub-grid variability and processes (e.g., Langevin equations and fluctuating Navier-Stokes equations). Solutions of such problems, or QoIs derived from them, are given in terms of their probability density functions (PDFs). The goal here is to learn the deterministic dynamics of either the PDF of a system state (e.g., the Fokker-Planck equation for a given Langevin equation~\citep{Risken1989Fokker}) or its statistical moments (e.g., a PDE describing the spatiotemporal evolution of the ensemble mean of the system state~\citep{winter2003moment}).
%Probabilistic models are essential in various fields of science and engineering, where they are used to make science-based predictions under epistemic and aleatory uncertainty. Their applications spans a wide range of fields from biology~\citep{ye-2019-quantification} and environmental forecasting~\citep{tartakovsky2013assessment} to robotics~\citep{thrun2002probabilistic} and econometrics~\citep{mcfadden1981econometric}.
Human (as opposed to machine) learning of such deterministic PDEs or their nonlocal (integro-differential) counterparts relies, e.g., on stochastic homogenization of the underlying stochastic models or on the method of distributions~\citep{tartakovsky-2016-method}. The latter provides a systematic way to derive deterministic PDF or CDF (cumulative distribution function) equations, regardless of whether the noise is white or colored~\citep{wang2013probability}. Stochastic computation via the method of distributions can be orders of magnitude faster than high-resolution Monte Carlo~\citep{alawadhi2018method, yang2019probabilistic}. 

While under certain conditions PDF equations can be exact, in general (e.g., when the noise is multiplicative and/or correlated) their derivation requires a closure approximation~\citep{alawadhi2018method, venturi-2013-exact, yang2019probabilistic}. Such closures are usually derived either through perturbation expansions in the (small) variances of the input parameters or by employing heuristic arguments. Both approaches require considerable field-specific knowledge and can introduce uncontrollable errors. We propose replace the human learning of PDF/CDF equations with a machine learning method to infer closure terms from data. It is based on sparse regression for discovering relevant terms in a differential equation~\citep{Brunton2016, Schaeffer2017, schmidt2009distilling}, although its goals are different. The data come from a relatively few Monte Carlo runs of the underlying differential equation with random inputs, rather than from elusive observational data. Our approach amounts to coarse-graining in probability space and is equally applicable to deterministic coarse-graining as well.

We posit that sparse regression for PDE learning is better suited for PDF/CDF equations than for general PDEs. First, random errors in data and/or random fluctuations in an underlying physical process undermine the method's ability to learn a governing equation~\citep{Rudy2019}; yet, their distributions might be easier to handle because of the smoothness of corresponding PDFs/CDFs~\citep{boso-2020-learning, boso2020data}. Second, the known properties of distributions and PDF/CDF equations significantly constrain the dictionary of possible terms, rendering the equation learning more tractable and truly physics-informed. For example, a PDF equation has to be conservative (i.e., has to conserve probability); and, according to the Pawula theorem~\cite[pp.~63-95]{Risken1989Fokker}, the Kramers-Moyal expansion (i.e., a Taylor-series expansion of a master equation) should stop at the first three terms to preserve a PDF's positivity (giving rise to the Fokker-Plank equation). Finally, PDF equations tend to be linear, even if the underlying physical law describing each realization is nonlinear~\citep{tartakovsky-2016-method}, which also limits the dictionary's size. Such considerations are, or should be, a key feature of \emph{physics-informed} machine learning. %Yet, they are often absent in the equation-learning efforts.
%This linearization is done by projection on a higher dimensional space.

Our strategy to learn PDF equations from noisy data is presented in \cref{sec:method}. A series of computational experiments in \cref{sec:results} is used to illustrate the robustness and accuracy of our approach. Main conclusions drawn from our study are summarized in \cref{sec:concl}.
%In this study, we emphasize the use of machine learning as part of a semi-analytical method to discover a PDF equation.

%%%%%%%%%%%%%%%%%%%%%%%%%%%%%%%%%%%%%%%%%%%%%%%%%%%%%%%%%%%%%%
\section{Autonomous learning of PDF equations and their closure approximations} % ...of PDF (coarse-grained) equations?
\label{sec:method}
%%%%%%%%%%%%%%%%%%%%%%%%%%%%%%%%%%%%%%%%%%%%%%%%%%%%%%%%%%%%%%

We start by formulating in \cref{sec:problem} a generic problem described by a nonlinear PDE with uncertain (random) parameters and/or driving forces. A deterministic equation for the PDF of its solution is formulated in \cref{sec:PDFeq}. 
%Our physics-informed sparse-regression strategy for identifying a linear differential operator in this PDF equation is presented in \cref{sec:regression}.
In \cref{sec:regression}, we present two sparse-regression strategies for discovery of PDF equations. These are referred to as direct equation learning (DEL) and constrained equation learning (CEL).

\subsection{Problem formulation}
\label{sec:problem} 

Consider a real-valued system state $u(\mathbf x, t): D \times \mathbb R^+ \to D_u$ that is defined on the $d$-dimensional spatial domain $D \subset \mathbb R^d$ and has a range $D_u \subset \mathbb R$. Its dynamics is described by a PDE 
\begin{equation}\label{geneqn}
    \frac{\partial u }{\partial t} + \mathcal N_\mathbf{x}(u;{\bm \lambda}_{\mathcal N}) = g(u; {\bm \lambda}_g), \qquad \mathbf x \in D, \quad t > 0,
\end{equation}
which is subject to a initial condition $u(\mathbf x, 0) = u_\text{in}(\mathbf x)$ and boundary conditions on the boundary $\partial D$ of $D$ (to be specific, and without loss of generality, we consider a Dirichlet condition $u(\mathbf x, t) = u_\text{b}(\mathbf x, t)$ for $\mathbf x \in \partial D$). The linear or nonlinear differential operator $\mathcal N_\mathbf{x}$ contains derivatives with respect to $\mathbf x$ and is parameterized by a set of coefficients $\bm \lambda_{\mathcal N}(\mathbf x, t)$. The source term $g(u)$, a real-valued smooth function of its argument, involves another set of parameters ${\bm \lambda}_g(\mathbf x,t)$. The system parameters $\bm \lambda = \{\bm \lambda_{\mathcal N}, {\bm \lambda}_g \}$ are uncertain and treated as random fields. They are characterized by a single-point joint PDF $f_{\bm \lambda}(\bm \Lambda; \mathbf x, t)$ and a two-point covariance function (a matrix) $\mathbf C_{\bm \lambda}(\mathbf x,t; \mathbf y,\tau)$, both of which are either inferred from data or provided by expert knowledge. The auxiliary functions $u_\text{in}(\mathbf x)$ and $u_\text{b}(\mathbf x, t)$ are also uncertain, being characterized by their respective single-point PDFs $f_{u_\text{in}}(U; \mathbf x)$ and $f_{u_\text{b}}(U; \mathbf x, t)$ and appropriate spatiotemporal auto-correlation functions.

%be a random field with a random sample space $\Omega$ s.t. $\omega \in \Omega$, and let $\bm \lambda(x, t; \omega)$ be a vector of parameters with a known distribution. 
%In what follows, the random field sample variable $\omega$ is omitted for simpler notation.

Uncertainty in the input parameters renders predictions of the system state $u(\mathbf x,t)$ uncertain (random) as well. Consequently, a solution to~\eqref{geneqn} is the PDF $f_u(U; \mathbf x, t)$ of $u(\mathbf x,t)$, whose mean $\mathbb E(u) \equiv \bar u(\mathbf x,t) = \int U f_u(U; \mathbf x, t) \text dU$ and variance $\sigma_u^2(\mathbf x,t) = \int U^2 f_u(U; \mathbf x, t) \text dU - \bar u^2$ serve as an unbiased prediction and a measure of predictive uncertainty, respectively. Here, the integration is over $D_u$.

Multiple uncertainty propagation tools can be used to estimate the PDF $f_u(U; \mathbf x, t)$. These include (multilevel) Monte Carlo simulations (e.g.,~\cite{taverniers-2020-estimation} and the references therein), which require one to draw multiple realizations of the inputs $\{\bm \lambda, u_\text{in}, u_\text{b} \}$ and solve~\eqref{geneqn} for each realization. This and other uncertainty quantification techniques are typically computationally expensive and provide little (if any) physical insight into either the expected (average) dynamics or the dynamics of the full PDF $f_u$. The method of distributions provides such an insight by yielding a deterministic PDE, which describes the spatiotemporal evolution of $f_u(U; \mathbf x, t)$.

\subsection{PDF equations}
\label{sec:PDFeq}

Regardless of whether the differential operator $\mathcal N_\mathbf{x}$ in~\eqref{geneqn} is linear or nonlinear, the PDF $f_u(U; \mathbf x, t)$ satisfies a $(d+1)$-dimensional linear PDE~\citep{tartakovsky-2016-method}
\begin{equation}\label{eq:PDF}
    \frac{\partial f_u}{\partial t} + \mathcal L_{\tilde{\mathbf x}} (f_u; \bm \beta) = 0, \qquad \tilde{\mathbf x} \equiv (\mathbf x,U) \in D \times D_u, \quad t > 0,
\end{equation}
with a set of coefficients $\bm \beta(\tilde{\mathbf x}, t)$. According to the Pawula theorem~\cite[pp.~63-95]{Risken1989Fokker}, the linear differential operator $\mathcal L_{\tilde{\mathbf x}}$ can include either first and second or infinite-order derivatives with respect to $\tilde{\mathbf x}$. Transition from~\eqref{geneqn} to~\eqref{eq:PDF} involves two steps: projection of the $d$-dimensional (linear or nonlinear) PDE~\eqref{geneqn} onto a $(d+1)$-dimensional manifold with the coordinate $\tilde{\mathbf x}$, and coarse-graining (stochastic averaging) of the resulting $(d+1)$-dimensional linear PDE with random inputs.\footnote{When the system parameters $\bm \lambda$ vary in space and/or time, PDF equations are typically space-time nonlocal~\cite{barajas2016probabilistic, maltba-2018-nonlocal}, i.e., integro-differential, and the derivation of~\eqref{eq:PDF} requires an additional localization step.} For first-order hyperbolic PDEs, this procedure can be exact when the system parameters $\bm \lambda$ are certain~\cite{alawadhi2018method} and requires closure approximations otherwise~\cite{boso-2014-cumulative}. It is always approximate when PDEs involved are parabolic~\cite{boso-2016-method} or elliptic~\cite{yang2019probabilistic}, in which case the meta-parameters $\bm \beta$ might depend on the moments of the PDF $f_u$ in a manner akin to the Boltzmann equation. Identification of the coefficients $\bm \beta(\tilde{\mathbf x}, t)$, some of which might turn out to be 0, is tantamount to physics-informed learning of PDF equations. 

When the system parameters $\bm \lambda$ are random constants---or when a space-time varying  parameter, e.g., random field $\lambda(\mathbf x)$, is represented via a truncated Karhunen-Lo\`eve expansion in terms of a finite number $N_\text{KL}$ of random variables $\lambda_1,\ldots,\lambda_{N_\text{KL}}$---the PDF equation~\eqref{eq:PDF} is approximate, but an equation for the joint PDF $f_{u \bm \lambda}(U,\bm \Lambda; \mathbf x,t)$ of the inputs $\bm \lambda$ and the output $u$,
\begin{equation}\label{pdfeqnlearn}
    \frac{\partial f_{u \bm \lambda}}{\partial t} + \hat{\mathcal L}_{\tilde{\mathbf x}} (f_{u \bm \lambda}; \hat{\bm \beta}) = 0, \qquad \tilde{\mathbf x} \equiv (\mathbf x,U) \in D \times D_u, \quad t > 0,
\end{equation}
is exact~\cite{venturi-2013-exact}. Similar to~\eqref{eq:PDF}, the differential operator $\hat{\mathcal L}_{\tilde{\mathbf x}}$ is linear and consists of up to second-order derivatives with respect to $\tilde{\mathbf x}$; its dependence on $\bm\Lambda$ is parametric, $\hat{\bm \beta} = \hat{\bm \beta}(\bm \Lambda, \mathbf x,t)$. Since the number of parameters in the set $\bm \lambda$ can be very large, one has to solve~\eqref{pdfeqnlearn} for multiple values of $\bm \Lambda$, which is computationally expensive. A workable alternative is to compute a PDF equation~\eqref{eq:PDF} for the marginal $f_u(U; \mathbf x,t)$ by integrating~\eqref{pdfeqnlearn} over $\bm \Lambda$. In general, this procedure requires a closure~\cite{venturi-2013-exact}.

\subsection{Physics-informed dictionaries}
\label{sec:regression}

%\textbf{[We shouldn't base our method on the Pawula theorem, to allow for integral features and potentially nonlinear PDF equations. We should use it as a constraint, just as we use the argument of conservation. We can also test if our approach works if it satisfies the theorem (although that's inconsistent with having integral terms). ]}
%And are we guaranteed that the equation of the marginal is described by master equation?

%Deep learning approaches to learning PDEs relies on a dictionary of terms that a differential operator can have. That is where a key advantage of our approach to learning the dynamics of $f_u$, as in~\eqref{eq:PDF}, rather than the underlying dynamics of $u$ in~\eqref{geneqn}, manifests itself . Specifically, the Pawula theorem provides both an exhaustive dictionary for PDF/CDF equations and the form of $\mathcal L_{\tilde x}$,
%

Traditional data assimilation approaches for parameter identification, and deep learning strategies for PDE learning, rely on \emph{a priori} knowledge of a dictionary of plausible terms in the differential operator. This is where a key advantage of learning the dynamics of $f_u(U;\mathbf x,t)$ in~\eqref{eq:PDF}, rather than the underlying dynamics of $u(\mathbf x,t)$ in~\eqref{geneqn}, manifests itself. Theoretical properties of PDF equations significantly constrain the membership in a dictionary, ensuring a faster and more accurate convergence to an optimal solution. We propose two strategies for discovering the PDF equation: DEL seeks to learn the full operator in equation \eqref{eq:PDF}, and CEL utilizes partial knowledge of the operator. This is illustrated in the following formulation of an optimization problem. 

% General formulation of optimization problem
Let $\hat{\mathbf f}_u \in \mathbb R^{M \times N \times P}$, with entries $\hat f_u^{ijk} \equiv f_u(U_i, \mathbf x_j, t_k)$ for $i \in [1, M]$, $j \in [1, N]$ and $k \in [1, P]$, be a (numerical) solution of~\eqref{eq:PDF}, at appropriate nodes of discretized $U \in D_u$, $\mathbf x \in D$, and $t \in [0, T]$, such that $U_i = U_0 + i\Delta U$, $\mathbf x_j = \mathbf x_0 + j\Delta \mathbf x$ and $t_k = t_0 + k\Delta t$. Our goal is to discover the differential operator
\begin{equation} \label{eq:operator-dictionary}
    \mathcal L_{\tilde{\mathbf x}} = \bm \beta(\tilde{\mathbf x}, t) \cdot \underbrace{ \left( 1, \frac{\partial }{\partial \tilde x_1}, \cdots, \frac{\partial }{\partial \tilde x_{d+1}}, \frac{\partial^2}{\partial \tilde x_1^2},  \frac{\partial^2}{\partial \tilde x_1 \partial \tilde x_2}, \cdots, \frac{\partial^2}{\partial \tilde x_{d+1}^2},  \cdots \right)^\top}_{\text{The dictionary $\mathcal H$ consisting of $Q$ members} },
\end{equation}
where $\bm \beta(\tilde{\mathbf x}, t) = (\beta_1,\ldots,\beta_Q)^\top \in \mathbb R^Q$ is the $Q$-dimensional vector of unknown (variable) coefficients. %that have $Q$ components, $\beta_q(U,x,t)$ with $q=1,\ldots,Q$, multiplying $Q$ derivative features. 
This is accomplished by minimizing the discretized residual 
\begin{align}\label{eq:residual}
\mathcal R_{ijk}(\bm \beta) = \frac{\partial \hat f^{ijk}_u}{\partial t} + \mathcal L_{\tilde{\mathbf  x}} (\hat f^{ijk}_u; \bm \beta),
\end{align}
for all grid points $(U_i, \mathbf x_j, t_k)$. The derivatives in~\eqref{eq:operator-dictionary} are approximated via finite differences, fast Fourier transforms, total variation regularized differentiation, etc., as discussed in~\cite{Brunton2016, Schaeffer2017}. Accordingly, the vector of optimal coefficients, $\check{\bm \beta}$, is found as a solution of the minimization problem
\begin{align}
\label{eq:generalopt}
    \check{\bm \beta}(U, \mathbf x, t) = \underset{\bm \beta(U, \mathbf x, t)}{\text{argmin}} \left\{%\int_0^T \int_{\mathbf y \in \mathcal D} 
     \frac{1}{MNP} \sum_{i=1}^M \sum_{j=1}^N \sum_{k=1}^P \mathcal R^2_{ijk}(\bm \beta) %d\mathbf y \text dt 
    + \gamma || \bm \beta ||_1^2 \right\}.
\end{align}
The $L_1$ norm, $\| \cdot \|_1$, is a regularization term that provides sparsification of the PDF equation, with $\gamma$ serving as a hyper-parameter coefficient. The residual, $\mathcal R_{ijk}$, represents a single training example indexed by the triplet $(i, j, k)$. In what follows, the subscript $\cdot_{ijk}$ is sometimes omitted to simplify the notation. Discovery of the full operator $\mathcal L_{\tilde{\mathbf x}}$, i.e., the solution of~\eqref{eq:operator-dictionary}--\eqref{eq:generalopt} is referred to as DEL.
%The triple sum can be replaced by a Frobenius norm $L_\text{F}$, $\| \cdot \|_\text{F}$ assuming the residual is a 3D tensor. But it makes more sense to express it this way to be consistent with the literature. Here, every (i, j, k) is a single training example, and the minimization is done by summing over all training examples.

% Motivation for constraining the hypothesis set
The challenge in making the optimization problem~\eqref{eq:generalopt} generalize to unseen space-time points is to identify a proper dictionary of derivatives in~\eqref{eq:operator-dictionary} that balances model complexity and predictability. On one hand, a larger hypothesis class $\mathcal H$ (here, parametrized by $Q$ coefficients $\beta_q(U, \mathbf x, t)$ with $q = 1,\ldots, Q$) has a higher chance of fitting the optimal operator $\mathcal L_{\tilde{\mathbf x}}$ that honors $\hat{\mathbf f}_u$. It does so by minimizing the bias at the cost of a higher variance. On the other hand, a smaller dictionary $\mathcal H$ discards hypotheses with large variance, automatically filtering out noise and outliers that prevent the model from generalizing.

Both features are often used in coordination to nudge the regression problem in the right direction. For instance, having variable instead of constant coefficients in \eqref{eq:generalopt} significantly increases the power of the model to describe simulation data. At the same time, the $L_1$ regularization favors the parsimony of (i.e., fewer terms in) the operator $\mathcal L_{\tilde{\mathbf x}}$; making the resulting PDF equation more interpretable and easier to manipulate analytically.
%bounds the hypothesis to sparse solutions with small coefficients; typically associated with the law of parsimony, making the equations more interpretable and easier to manipulate analytically.

The construction of the dictionary in~\eqref{eq:operator-dictionary} and, hence, of the residual $\mathcal R(\bm \beta)$ is guided by the following considerations. 
%In general, learning equations of physical systems, the hypothesis class can be significantly constrained by \emph{a priori} knowledge of deterministic physical properties (such as conservation of mass, energy and momentum) and mathematical properties of differential equations (such as non-negativity, linearity etc.). These constraints can be factored in the form of the residual $\mathcal R(\bm \beta)$. We elaborate on this point in this section.
% Pawula theorem constrains the number of derivatives
First, if the random state variable $u(\mathbf x, t)$ is represented by a master equation, the Pawula theorem provides an exhaustive dictionary for PDF/CDF equations, i.e., specifies the form of $\mathcal L_{\tilde{\mathbf x}}$. It states that a truncated Taylor expansion of the master equation (i.e., the Kramers-Moyal expansion) must contain no higher than second-order derivatives for the function $f_u$ to be interpretable as a probability density; otherwise, it can become negative. Consequently, if we restrict our discovery to local PDEs, i.e., ignore the possibility of $f_u$ satisfying integro-differential equations or PDEs with fractional derivatives, then the dictionary containing first and second order derivativex terms in~\eqref{eq:operator-dictionary} is complete. %Otherwise, a solution of the discovered PDE would violate the axioms of probability (non-negativity and integration to one). Therefore, the Pawula theorem restricts the number of derivative features to those explicitly shown in ~\eqref{eq:operator-dictionary} (i.e. $Q=6$).

% PDF equations can be written (and discretized) in conservative form
Second, the learned PDE for $f_u(U, \mathbf x, t)$ has to conserve probability, $\int f_u \text dU = 1$ for all $(\mathbf x, t) \in D \times [0, T]$, i.e., the differential operator in~\eqref{eq:operator-dictionary} must be of the form $\bar{\mathcal L}_{\tilde{\mathbf x}} = \nabla_{\tilde{\mathbf x}} \cdot (\bar{\bm \beta} \nabla_{\tilde{\mathbf x}})$,
%\begin{equation}\label{eq:conservative-form}
%\mathcal R(\bm \beta) = \frac{\partial \hat f_u}{\partial t} + \nabla_{\tilde x} \bar{\mathcal  L}_{\tilde{\mathbf x}}  (\hat f_u; \bar{\bm \beta} ), 
%\end{equation}
%
where $\bar \cdot$ designates operators, and their coefficients, in the conservative form of the PDF equation. Accordingly, $\bar{\mathcal L}_{\tilde{\mathbf x}}(\cdot;\bar{\bm \beta})$ is a subset of its non-conservative counterpart $\mathcal L_{\tilde{\mathbf x}}(\cdot; \bm \beta)$ in \eqref{eq:PDF}. 
The conservative form does not only constrain the form of the operator, but also facilitates its numerical approximation. For example, a conservation law can be discretized using a finite volume scheme ensuring that the learnt solution conserves probability.

% Analytical methods like the method of distributions discover part of the operator
In a typical coarse-graining procedure, only a fraction of the terms in a PDF/CDF equation (i.e, in the dictionary $\mathcal H$), those we refer to as closures, are unknown~\cite{tartakovsky-2016-method} and need to be learned from data. For example, an ensemble mean $\langle I O \rangle$ of two random fields, the model input $I$ and (a derivative of) the model output $O$ is written as $\langle I O \rangle = \langle I \rangle \langle O \rangle +  \langle I' O' \rangle$, where the prime $^\prime$ indicates zero-mean fluctuations about the respective means.  The first term in this sum is a known term in a coarse-grained PDE, while the second requires a closure approximation, i.e., needs to be discovered. When applied to~\eqref{geneqn}, the method of distributions~\cite{tartakovsky-2016-method} leads to an operator decomposition $\mathcal L_{\tilde{\mathbf x}} = \mathcal K_{\tilde{\mathbf x}} + \mathcal C_{\tilde{\mathbf x}}$, where $\mathcal K_{\tilde{\mathbf x}}$ is a known differential operator and the unknown operator $\mathcal C_{\tilde{\mathbf x}}$ contains the closure terms to be learned.  With this decomposition, the discretized residual~\eqref{eq:residual} takes the form
\begin{equation}\label{eq:closure-residual}
    \mathcal R_{ijk}(\bm \beta) = \frac{\partial \hat f_u^{ijk}}{\partial t} + \mathcal  K_{\tilde{\mathbf x}}(\hat f_u^{ijk}; \bm \eta) + \mathcal  C_{\tilde{\mathbf x}} (\hat f_u^{ijk}; \bm \beta) ,
\end{equation}
%%% Conservative form
% \begin{equation}
%     \mathcal R(\bm \beta) = \frac{\partial \hat f_u}{\partial t} + \nabla_{\tilde x} \cdot \left[ \bar{\mathcal  K}_{\tilde x}(\hat f_u; \bar{\bm \eta}) + \bar{\mathcal  C}_{\tilde x} (\hat f_u; \bar{\bm \beta}) \right],
% \end{equation}
with known coefficients $\bm \eta$ and unknown coefficients $\bm \beta$, which are a subset of their counterparts in~\eqref{eq:residual}. Minimization of the residual~\eqref{eq:closure-residual} lies at the heart of CEL. We posit that CEL provides a proper framework for physics-informed  equation discovery, in which physics informs the construction of the operator $\mathcal K_{\tilde{\mathbf x}}$ and observational/simulated data are used to infer the unknown closure operator $\mathcal C_{\tilde{\mathbf x}}$.  In general, there are many more ways to constrain the dictionary $\mathcal H$ based on physical and mathematical properties of the differential equations one aims to learn. Depending on the problem, the scientific literature is full of versatile physical constraints that can and should improve the discovery of the equations.

\begin{remark}
While generalization is what all human and machine learning aims to achieve, experience shows that the set over which a model generalizes is always bounded. This is why it is important to keep the human in the loop of discovering ever more generalizing models by learning interpretable models. With that purpose in mind, while deep learning techniques are good at fitting nonlinear functions, learning equations by sparse regression provides a better collaborative framework between the scientist and the machine.
\end{remark}

\subsection{Numerical implementation}

Since the coefficients $\bm\beta(U, \mathbf x, t)$ are functions of $(d+2)$ arguments, a numerical solution of the optimization problem in~\eqref{eq:generalopt} might be prohibitively expensive. For example, a simulated annealing strategy (e.g.,~\cite{boso2018information} and the references therein) calls for discretizing the coefficients $\bm \beta^{ijk} = \bm \beta(U_i, \mathbf x_j, t_k)$ at  the grid points $(U_i, \mathbf x_j, t_k)$ at which the solution $\hat f_u^{ijk}$ is defined and optimizing over $\bm \beta^{ijk}$. With $Q$ features in the dictionary $\mathcal H$, this strategy yields $Q \times M \times N \times P$ unknown coefficients $\beta_q^{ijk}$ and an optimization problem with complexity $\mathcal O(QM^3)$, where typically $M \approx 10^3$. Solving such a  high-dimensional problem requires adequate computational resources, i.e., multithreading on GPUs, proper memory allocation, etc. It can be implemented by stacking the minimization problems over all grid points in one large matrix, as done in~\cite{Rudy2019} for learning parametric PDEs.

A more efficient approach is to represent the variable coefficients $\beta_q(U, \mathbf x, t)$ via a series of orthogonal polynomial basis functions (e.g., Chebyshev polynomials), $\psi_r(\cdot)$, such that
\begin{equation}\label{eq:var-coef}
    \beta_q(U, \mathbf x, t) = \sum_r^R\sum_s^S\sum_w^W \alpha^{rsw}_q \psi_r(U) \psi_s(\mathbf x) \psi_w(t), \qquad q = 1,\ldots,Q, 
\end{equation}
%  \psi_i(U) \psi_j(x) \psi_k(t)
where $\alpha^{rsw}_q \in \mathbb R$ denote the $d_\text{pol} = RSW$ coefficients in the polynomial representation of $\bm \beta_q$. With this approximation, the minimization problem~\eqref{eq:generalopt} is solved over the unknown coefficients $\bm \alpha^{rsw} = (\alpha^{rsw}_1, \ldots, \alpha^{rsw}_Q) \in \mathbb R^Q$. 
%In addition, $\psi_r(\cdot)$ is a basis function such that $\langle \psi_r, \psi_s \rangle = 0$ over a given interval; although orthogonality is not required for the optimization problem to work. 
For $d_\text{coef} = Q d_\text{pol}$ unknown coefficients $\beta_q^{ijk}$, the optimization dimension is now of order $\mathcal O(QR^3)$, where typically $R \lesssim 10$. This dimension is many orders of magnitude smaller than the brute force parametric optimization in~\cite{Rudy2019}, so that the resulting optimization problem can be solved on a personal computer.

\begin{figure}[htbp]
\begin{center}
\includegraphics[width=10cm]{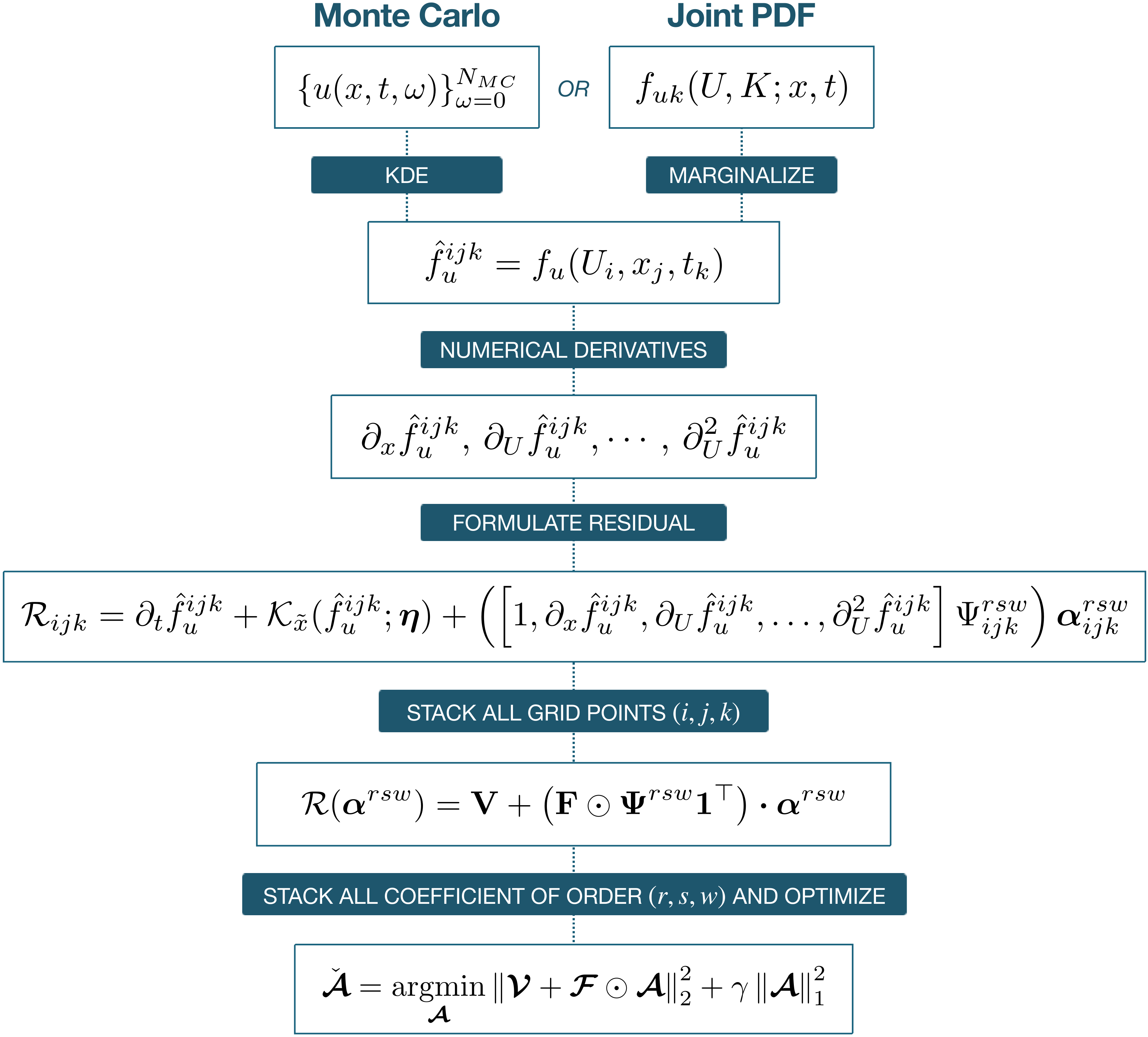}
\caption{A diagram of the algorithm for learning PDF equations from Monte Carlo simulations.}
\label{adv-react-sol}
\end{center}
\end{figure}

% Numerical linear algebra (matrix formulation details)
Given the data matrix $\hat{\mathbf f}_u \in \mathbb R^{M  \times N  \times P}$ and its numerical derivatives with respect to $U$, $\mathbf x$ and $t$ from the dictionary~\eqref{eq:operator-dictionary}, we build the derivative feature matrix
\begin{align}
\mathbf F = 
  \begin{bmatrix}
    1 & \partial_{x_1} \hat f_u^{111} & \cdots &  \partial_{x_d} \hat f_u^{111} &  \partial_U \hat f_u^{111} & \cdots & \partial_U^2 \hat f_u^{111} \\
    1 & \partial_{x_1} \hat f_u^{211} & \cdots &  \partial_{x_d} \hat f_u^{211} &  \partial_U \hat f_u^{211} & \cdots & \partial_U^2 \hat f_u^{211} \\
    \vdots & \vdots & \vdots & \ddots & \vdots \\
    1 &\partial_{x_1} \hat f_u^{MNP} & \cdots &  \partial_{x_d} \hat f_u^{MNP} &  \partial_U \hat f_u^{MNP} & \cdots & \partial_U^2 \hat f_u^{MNP} \\
  \end{bmatrix}
  \in \mathbb R^{d_\text{dis} \times Q}, \quad d_\text{dis} = MNP;
\end{align}
and its corresponding label vector (i.e., the known part of the PDF equation); e.g., based on the CEL formulation of the residual in~\eqref{eq:closure-residual},
\begin{align}
\mathbf V = 
  \begin{bmatrix}
    \partial_t \hat f_u^{111} + \mathcal K_{\tilde{\mathbf x}}(\hat f_u^{111}; \bm \eta) \\
    \partial_t \hat f_u^{211} + \mathcal K_{\tilde{\mathbf x}}(\hat f_u^{211}; \bm \eta) \\
    \vdots \\
     \partial_t \hat f_u^{MNP} + \mathcal K_{\tilde{\mathbf x}}(\hat f_u^{MNP}; \bm \eta)
  \end{bmatrix}
  \in \mathbb R^{d_\text{dis}}.
\end{align}
For variable coefficients $\bm \beta(U, \mathbf x, t)$, we define the vector $\mathbf \Psi^{rsw} \in \mathbb R^{d_\text{dis}}$ whose elements $\Psi^{rsw}_{ijk} \equiv \psi_r(U_i) \psi_s(\mathbf x_j)$ $\psi_w(t_k)$ correspond to the grid-point elements in the columns of $\mathbf F$ and $\mathbf V$.
For every polynomial coefficient vector $\mathbf \Psi^{rsw}$, the matrix form of the residual in \eqref{eq:closure-residual} becomes
\begin{equation}\label{eq:matrix-residual}
    \mathcal R(\bm \alpha^{rsw}) = \mathbf V + (\mathbf F \odot \bm \Psi^{rsw} \mathbf 1^\top) \bm \alpha^{rsw}, 
\end{equation}
%\mathcal R_{ijk} = \partial_t \hat f_u^{ijk} + \mathcal K_{\tilde x}(\hat f_u^{ijk}; \boldsymbol \eta) + \left( \left[1, \partial_x \hat f_u^{ijk}, \partial_U \hat f_u^{ijk}, \ldots, \partial^2_U \hat f_u^{ijk} \right] \Psi_{ijk}^{rsw} \right) \boldsymbol \alpha_{ijk}^{rsw}
where $\odot$ is the Hadamard (element-wise) product, $\mathbf 1 \in \mathbb R^Q$ is a vector of ones, such that the outer product $\mathbf \Psi^{rsw} \mathbf 1^\top$ broadcasts the variable coefficient vector $\mathbf \Psi^{rsw}$ into $Q$ identical columns. Let us introduce matrices
\begin{align}
\bm{\mathcal V} = 
    \begin{bmatrix}
     \mathbf V \\ 
    \mathbf V  \\
    \vdots \\
    \mathbf V 
  \end{bmatrix} \in \mathbb R^{d_\text{tot} },
\qquad
\bm{\mathcal F} = 
    \begin{bmatrix}
    \mathbf F \odot \mathbf \Psi^{111} \mathbf 1^\top \\ 
    \mathbf F \odot \mathbf \Psi^{211} \mathbf 1^\top\\
    \vdots \\
     \mathbf F \odot \mathbf \Psi^{RSW} \mathbf 1^\top 
  \end{bmatrix} \in \mathbb R^{d_\text{tot} \times Q },
\qquad
\bm{\mathcal A} =
  \begin{bmatrix}
   \bm \alpha^{111} \\ 
   \bm \alpha^{211} \\
    \vdots \\
   \bm \alpha^{RSW}
  \end{bmatrix} \in \mathbb R^{d_\text{coef} }, 
\end{align}
where $d_\text{tot} = d_\text{dis} d_\text{pol}$. Then, minimization of the residual in~\eqref{eq:matrix-residual} over all variable coefficients leads to the  optimization problem
%\[
%\check{\mathbf A} \equiv
%  \begin{bmatrix}
%   \check{\bm \alpha}^{111} \\ 
%   \check{\bm \alpha}^{211} \\
%    \vdots \\
%   \check{\bm \alpha}^{RSW}
%  \end{bmatrix}
%  = 
%  \underset{\mathbf A}{\text{argmin}}
%   \left\|
%    \begin{bmatrix}
%     \mathbf V \\ 
%    \mathbf V  \\
%    \vdots \\
%    \mathbf V 
%  \end{bmatrix}
%  +
%  {
%  %\underbrace{
%    \begin{bmatrix}
%    \mathbf F \odot \mathbf \Psi^{111} \mathbf 1^\top \\ 
%    \mathbf F \odot \mathbf \Psi^{211} \mathbf 1^\top\\
%    \vdots \\
%     \mathbf F \odot \mathbf \Psi^{RSW} \mathbf 1^\top 
%  \end{bmatrix}
%  %}_{MNPRSW\times Q}
%  }
%  \odot
%  \begin{bmatrix}
%    \bm \alpha^{111} \\ 
%   \bm \alpha^{211} \\
%    \vdots \\
%   \bm \alpha^{RSW}
%  \end{bmatrix}
%   \right\|^2_2
%  +
%  \gamma \left\| \bm A \right\|^2_1
%\]
\begin{equation}\label{eq:matrix-opt}
    \check{\bm{\mathcal{A}}} = \underset{\bm{\mathcal A}}{\text{argmin}} \left\| \bm{\mathcal V} + \bm{\mathcal F} \odot \bm{\mathcal A} \right\|^2_2 + \gamma \left\| \bm{\mathcal A} \right\|^2_1,
\end{equation}
where %$\bm A \in \mathbb R^{QRSW}$,  $\bm V \in \mathbb R^{MNPRSW}$ is a stack of $R\times S\times W$ vectors $\mathbf V$, $\bm F \in \mathbb R^{MNPRSW \times Q}$, and 
$\left\| \cdot \right\|_2$ denoting the $L_2$ norm. A schematic representation of the resulting algorithm is shown in Figure~\ref{adv-react-sol}.
% NOTE: The Hadamard product $\odot$ can be replaced by a matrix multiplication with diagonal entries, and the transpose by a dot product.
% Residual is of the form

% Algorithm details 
Following~\cite{Brunton2016}, our algorithm combines LASSO~\cite{tibshirani1996regression}, i.e,  $L_1$ regularization, with recursive feature elimination (RFE), which sequentially eliminates derivative features with small coefficients based on a tunable threshold at every iteration. This means that our algorithm has two hyper-parameters, $\gamma$ and the RFE threshold, which are chosen based on the test set error (rather than being part of the optimization variable $\mathbf A$) and a desired sparsity (i.e., a variance-bias balance). For this purpose, we test a few cross-validation algorithms for parameter estimation from Python's \texttt{scikit-learn} package \cite{scikit-learn}. These algorithms, which rely on grid search to find the optimal regularization hyper-parameter $\gamma$, are \texttt{LassoCV} (an $n$-fold cross-validation set on each iteration), \texttt{LassoLarsCV} (an additional least-angle regression model), and \texttt{LassoLarsIC} (the Akaike or Bayes information criterion as an optimization variable over $\gamma$). They give very similar results when the optimal solution is in the vicinity of the hypothesis class, but might differ significantly when the solution is far from optimal. In general, the choice of the algorithm depends on whether one favors more sparsity or accuracy.

% MC Processing
For $N_\text{MC}$ realizations of the random inputs,~\eqref{geneqn} is solved $N_\text{MC}$ times on the discretized space-time domain $D \times [0, T]$, yielding $N_\text{MC}$ solutions $u(\mathbf x, t)$. These Monte Carlo results are then post-processed, e.g., with a Gaussian kernel density estimator (KDE) used in this study, to obtain the single-point PDF $f_u(U;\mathbf x,t)$ on the discretized domain $D_u \times D \times [0, T]$. The KDE bandwidth is estimated for every grid point in $D \times [0, T]$ using Scott's normal reference rule $h = 3.49\, \sigma \,N_\text{MC}^{-1/3}$ \cite{scott1979optimal}, where $\sigma$ is the standard deviation of the data. The effect of the bandwidth on the solution is discussed in appendix \ref{app:hypertuning}. Both the kernel type and the bandwidth are added hyper-parameters that can be optimized.

% Threshold over f_t for reducing dimensionality
The matrices $\bm{\mathcal V}$ and $\bm{\mathcal F}$ in \eqref{eq:matrix-opt} can be very large, depending on the selected order of the polynomials ($R$, $S$ and $W$). We assume the coefficients to be time-independent, $\bm \beta = \bm \beta(U, \mathbf x)$, so that $W=1$. This makes the resulting optimization problems  numerically tractable on a personal computer. To increase the computational efficiency, we exclude grid points on which the labels, e.g., $\partial_t f_u(U; \mathbf x, t)$, remain close to zero during the entire simulation. This sampling method leads to a significant reduction in computational cost (around a four-fold reduction in matrix size), especially in the case of a PDF that remains unchanged (equal zero) on the majority of the infinite domain. 
% The effect of this reduction on the results is shown in appendix (can be added if needed).

% Training-Test Set size
To evaluate the generalization power of the method, we test its extrapolation power in time by fitting the hypothesis on the first $80\%$ of the time horizon $T$, i.e., on the domain $\mathcal D_\text{train} = D_u \times D \times [0, 0.8T]$, and testing it on the remaining $20\%$ of the simulation, i.e., on $\mathcal D_\text{test} = D_u \times \mathcal D \times [0.8T, T]$).

% Different Lasso algorithms used from python: LassoCV, LassoLarsCV, LassoLarsIC, LarsCV.

\section{Results}
\label{sec:results}

We validate our approach on a set of nonlinear problems with uncertain initial conditions and parameters.
In these experiments, we use the method of distributions \cite{tartakovsky-2016-method} to map the PDE~\eqref{geneqn} for the random field $u(\mathbf x, t)$ onto either closed or unclosed PDEs of the marginal PDF $f_u(U; \mathbf x, t)$. This illustrates the difficulties associated with analytical derivation of a PDF/CDF equation, and shows how our data-driven approach to PDE discover ameliorates them.

% \textcolor{red}{Does that fit in the intro better?}
%The overarching goal is to find an equation for $f_u(U; x,t)$ in the form \eqref{eq:PDF} from a single initial distribution, a single-point PDF of $u(x,t)$, aimed to generalize over space and time as well as all potential initial and boundary conditions. Experiments on a few test cases will prove the extrapolation power of the method. The purpose of finding an equation is to be used for computing the evolution of the probability distribution, in particular the mean solution, $\bar u(x,t) = \int U f_u(U; x,t) \text dU$ and/or other ensemble moments of the systems' behavior with a desired probability $\mathbb P[u(x,t) \le U]$.

\subsection{Nonlinear advection-reaction PDE with additive noise}
\label{sec:adv-react}

This experiment, in which the derivation of a PDF equation is exact, serves to test the method's accuracy in  reconstruction of a PDF equation from $N_\text{MC}$ Monte Carlo runs.  Let $u(x, t)$ be a real-valued state variable, whose dynamics is governed by
\begin{equation}\label{eq:adv-react}
    \frac{\partial u}{\partial t} + k \frac{\partial u}{\partial x} = r g(u), \qquad x \in \mathbb R, \quad t \in \mathbb R^+
\end{equation}
where $k \in \mathbb R^+$ and $r \in \mathbb R^+$ are deterministic advection and reaction rate constant, respectively. The initial condition $u(x, 0) = u_0(x)$ is a random field with compact support in $\mathbb R$; it is characterized by a single-point PDF $f_{u_0}(U; x)$ and a two-point correlation function $\rho_{u_0}(x, y)$ specified for any two points $x, y \in \mathbb R$. The nonlinearity $g(u)$ is such that for any realization of $u_0(x)$ a solution of this problem, $u(x,t)$, is almost surely smooth. The PDF $f_u(U;x,t)$ satisfies exactly a PDE (\ref{pdfderivation})
\begin{equation}\label{eq:pdf-adv-react}
    \frac{\partial f_u}{\partial t} + k \frac{\partial f_u}{\partial x} + r \frac{\partial g(U)f_u}{\partial U} = 0,
\end{equation}
subject to the initial condition $f_u(U;x,0) = f_{u_0}(U; x)$. 

For the nonlinear source $g(u) = u^2$ used in this experiment, the analytical solution of~\eqref{eq:adv-react} is $u(x, t) = [1/u_0(x - kt) - rt]^{-1}$. Uncertainty in the initial state, $u_0(x) = a \exp{[- (x - \mu)^2/(2\sigma^2)]}$, is incapsulated in the real constants $a$, $\mu$, and $\sigma$. These parameters are sampled from independent Gaussian distributions, $a \sim \mathcal N(\eta_a, \xi_a)$, $\mu \sim \mathcal N(\eta_\mu, \xi_\mu)$, $\sigma \sim \mathcal N(\eta_\sigma, \xi_\sigma)$. The means and variances in these distributions are chosen to ensure that $u_0(x)$ almost surely has a compact support, $u(x\rightarrow \pm \infty, t) = 0$, which ensures integrability of $u(x, \cdot)$ on $\mathbb R$. We set $k = 1$, $r=1$, $T = 0.5$, $\Delta t = 0.0085$, $x \in [-2.0, 3.0]$, $\Delta x = 0.0218$, $\Delta U = 0.0225$, $\eta_a = 0.8$, $\xi_a = 0.1$, $\eta_\mu = 0.5$, $\xi_\mu = 0.1$, $\eta_\sigma = 0.45$, $\xi_\sigma = 0.03$, and polynomial coefficients of order $M=3$ and $N=3$.

\begin{figure}[htbp]
\begin{center}
\includegraphics[width=15cm]{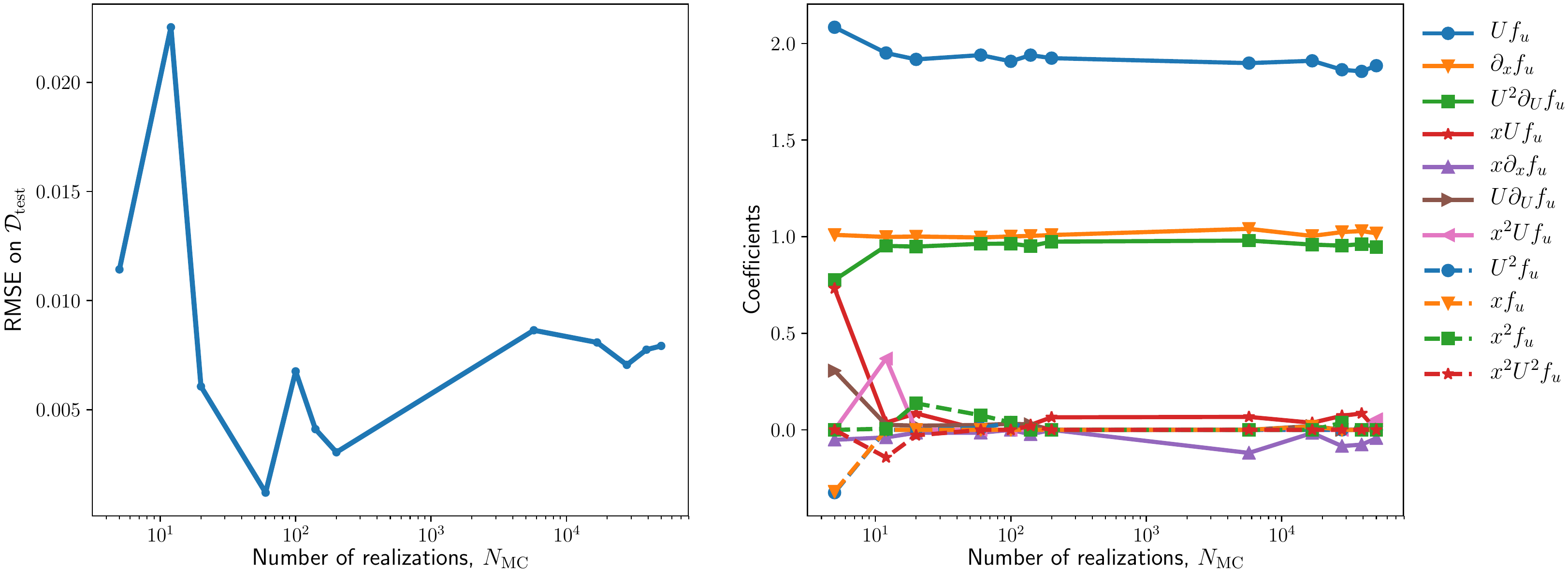}
\caption{Error in estimation of the PDF $f_u$ on $\mathcal D_\text{test}$ (left) and the coefficients in the discovered PDF equation~\eqref{eq:learnt-adv-react} (right) as function of the number of Monte Carlo realizations $N_\text{MC}$, without recursive feature elimination (RFE). 
}
\label{fig:adv-react-mc}
\end{center}
\end{figure}

We use the grid search algorithm \texttt{LassoCV} to find $\gamma = 0.0004$ that minimizes the test-set error, while seeking a sparse solution tunable by the RFE threshold. This direct equation learning (DEL) procedure leads to a PDE,
\begin{equation}\label{eq:learnt-adv-react}
    \frac{\partial \hat f_u}{\partial t} + \mathbf{0.996} \,\, \frac{\partial \hat f_u}{\partial x} + \mathbf{0.955} \,\, U^2 \frac{\partial \hat f_u}{\partial U} + \mathbf{2.06} \,\, U \frac{\partial \hat f_u}{\partial U} = 0,
\end{equation}
which demonstrates our method's ability to identify the relevant derivatives and their coefficients in equation~\eqref{eq:pdf-adv-react} with $g(U) \equiv U^2$, eliminating all the remaining features in the dictionary $\mathcal H$; the original coefficients $k=1$ and $r = 1$ are estimated with less than $5\%$ error. 
In the absence of recursive feature elimination, the algorithm yields 11 non-zero terms (Fig.~\ref{fig:adv-react-mc}), highlighting the importance of using RFE sparsification in addition to $L_1$ regularization.
%This test highlights the importance of sparsification, e.g., via RFE: the parameter inference via minimization of the test set error \textbf{[Eq. number]}, without sparsification ($\gamma = 0$) \textbf{[Right?]}, yields around \textbf{[What does ``around'' mean? -- it could be a bit different depending on other optimization parameters]} 10 non-zero terms (Fig.~\ref{fig:adv-react-mc}). 
This is due to the variance-bias trade-off discussed in section~\ref{sec:regression}.

The amount and quality of simulated data are characterized by two hyper-parameters: the number of Monte Carlo runs, $N_\text{MC}$, and the mesh size, $\Delta = \max \{\Delta U, \Delta x, \Delta t \}$. Figure~\ref{fig:adv-react-mc} reveals that both the values of the coefficients $\bm\beta$ in the PDF equation~\eqref{eq:learnt-adv-react} and the root mean square error (RMSE) its solution $\hat f_u$ in the extrapolation mode are relatively insensitive to $N_\text{MC}$ for around $N_\text{MC} > 20$ realizations. 
This means that in this particular problem, the required number of Monte Carlo simulations is very small. But this is not always the case, as will be shown in section~\ref{sec:burgers}.
The average RMSE is of order $\mathcal O(\Delta^{2})$, where $\Delta \approx 0.02$. This error is equivalent to a numerical scheme's approximation error (a truncation error of the relevant Taylor expansion). The second-order error observed here is due to the use of a first-order finite difference scheme to create the derivative features. Higher-order accuracy can be obtained by using a more accurate numerical method, e.g.,  FFT, for calculating the derivatives.

\begin{remark}
A solution $u(x,t)$ to~\eqref{eq:adv-react} can be (nearly) deterministic in a part of the space-time domain $\mathcal S \in D \times [0,T]$, e.g., when $u(x,t)$ has a compact support; in this experiment the size of $\mathcal S$ is controlled by the support of the initial state $u_0(x)$ which is advected by~\eqref{eq:adv-react} throughout the space-time domain $\mathbb R \times [0,T]$. This situation complicates the implementation of KDE and numerical differentiation, because the resulting PDF $f_u(U;x,t)$ is (close) to the Dirac delta function $\delta(\cdot)$; in this experiment, $f_u(U;x,t) \sim \delta(U)$ for $(x,t) \in \mathcal S$, as shown in Figure~\ref{fig:adv-react-boundary} for space-time points $(x=2.03,t)$ with small $t$. %Specifically, a deterministic $u(x, t)$ for a given point $(x, t)$ corresponds to a Dirac delta function $\delta(\cdot)$ in the PDF $f_u(U;x,t)$. In this experiment, the compact support of the solution $u(\pm \infty, t) = 0$ gives rise to its corresponding PDF solution $f_u(U; \pm \infty, t) = \delta(U)$, which cannot be eliminated from the simulation domain. 
Consequently, a numerical implementation of our algorithm must provide an adequate approximation of the delta function and be able to handle sharp gradients with respect to $U$ in the neighborhood of $U = 0$. (We found that rejecting data points near $u(x, t) = 0$ from KDE leads to a poor MC approximation of $f_u(U;\cdot)$ and its derivatives, and to the discovery of an incorrect PDF equations on $\mathcal D_\text{train}$.)
We address this issue by adding small perturbations $\xi$ to the initial state $u_0(x)$, i.e., by generating the training data from~\eqref{eq:adv-react} subject to the initial condition $u^\text{m}_0(x) = \xi + u_0(x)$, where the random variable $\xi$ has the exponential PDF, $f_\xi(s) = \lambda \exp(-\lambda s)$ for $s \ge 0$ and $= 0$ for $s < 0$, with $\lambda \gg 1$ (in our experiments, $\lambda =10$).\footnote{The choice of an exponential distribution ensures that $f_u(U; x, t) = 0$ for $U<0$, thus honoring the physical meaning of the random variable $u(x, t)$, e.g., solute concentration, that must stay positive throughout the simulation.} Another alternative is to omit training data from the simulation domain where the PDF has sharp profiles. In this case, the data in the domain $D^o_u \in [0, s |D_u |]$, with $s \in [0, 1]$, are excluded from the training set $\mathcal D_\text{train}$ (Fig.~\ref{fig:adv-react-boundary}b). Other strategies, which we defer for follow-up studies, include the discovery of PDF/CDF equations in the frequency domain.

\begin{figure}[htbp]
\begin{center}
\includegraphics[width=15cm]{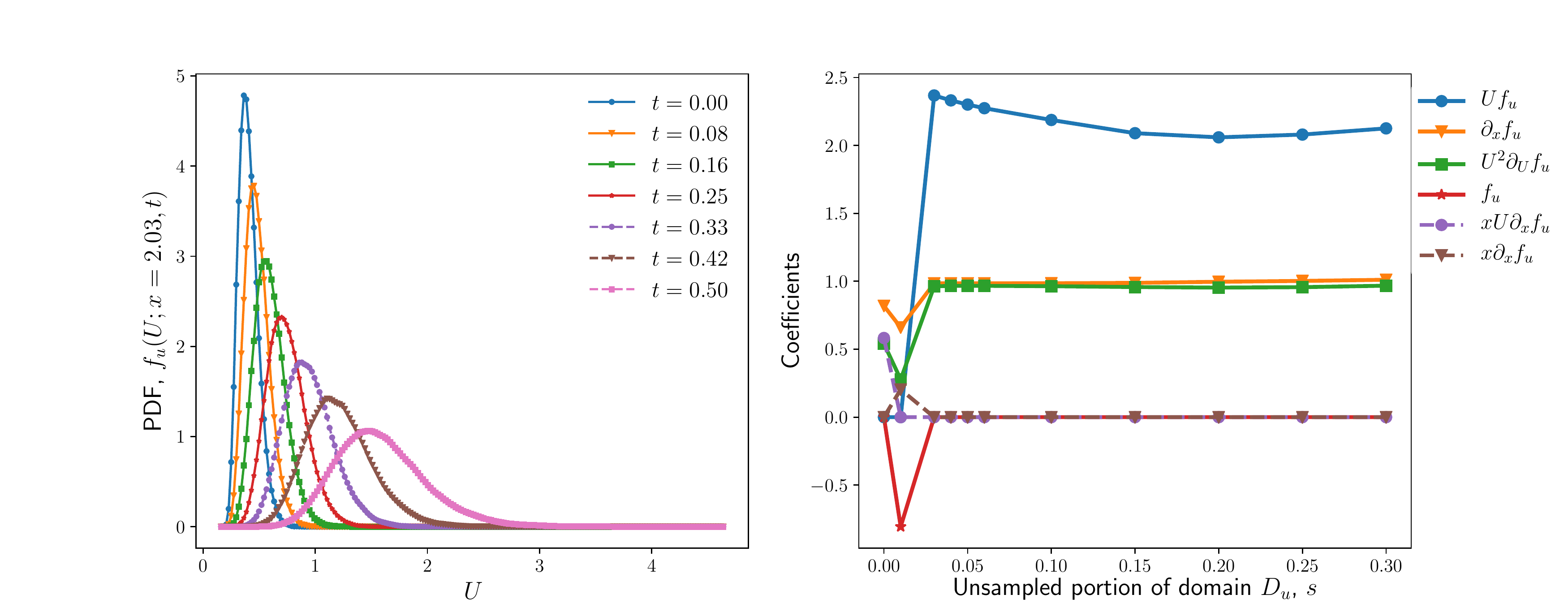}
\caption{  PDF $f_u(U;x=2.03,t)$ (left) and the coefficient values in the discovered PDF equation using DEL (right) The effect of omitting training samples from the $U$ domain $D^o_u \in [0, s|D_u |]$, with $s \in [0, 1]$, where the sharp PDF profiles complicate numerical differentiation. An RFE threshold of $0.1$ is used.}
\label{fig:adv-react-boundary}
\end{center}
\end{figure}

\end{remark}

%%%%%%%%%%%%%%%%%%%%%%%%%%%%%%%%%%%%%%%%%%%%%%%%%
\section{Conclusions and Discussion}
\label{sec:concl}

In general, coefficients $\beta_i(x, t)$ are functions of space and time. They are accordingly very high dimensional if simply discretized in $x$ and $t$.
One way to approximate variable coefficients is by using a polynomial expansion of the form $\beta_i(x, t) = \sum_j^n \alpha_{ij} \psi_j(x, t)$, where $\{\psi_j(x, t)\}_{j=1}^n$ are polynomial basis functions. 
We would now seek to learn $n$ constants $\alpha_j$ for each differential operator. 
In this case, we have $\mathcal O(nM)$ unknowns where $M$ is the size of the feature space (i.e. dictionary of derivative terms).
However, it is also possible to model $\beta _i(x, t)$ as a nonlinear function that can only be learnt numerically like a neural network. In this case, $n$ networks would be used with inputs $(x, t)$ and output $\beta_i(x, t)$.

In this study, we proposed a new method for learning equations for probability distribution function from abundant simulation equations.
The results show a promising direction for learning coarse-grained equations in general.
The similarity between the bias-variance trade-off and dimensional analysis demonstrates a promising direction for analyzing and deriving PDEs numerically.
Future work will explore the possibility of using nonlocal and variable coefficient features to improve accuracy and generality.

\appendix
%%%%%%%%%%%%%%%%%%%%%%%%%%%%%%%%%%%%%%%%%%%%%%%%%%%%%
%%%%%%%%%%%%%%%%%%%%%%%%%%%%%%%%%%%%%%%%%%%%%%%%%%%%%
\section{Derivation of the PDF Equation}
\label{pdfderivation}
%%%%%%%%%%%%%%%%%%%%%%%%%%%%%%%%%%%%%%%%%%%%%%%%%%%%%

Consider a generalized function 
\begin{equation}
\pi_u(U-u) \equiv \delta(U - u(x, t)),
\end{equation}
where $\delta(\cdot)$ is the Dirac delta function. If the random variable $u$ at any space-time point $(x,t)$ has a PDF $f_u(U;x,t)$, then, by definition of the ensemble average $\mathbb E[\cdot]$, 
\begin{align}\label{eq:avpi}
\begin{split}
\mathbb E[\pi_u(U-u)] & = \int_{-\infty}^{+\infty} \pi_u(U - \tilde U) f_u(\tilde U; x, t) \text d \tilde U \\
		& = \int_{-\infty}^{+\infty} \delta(U - \tilde U) f_u(\tilde U; x, t) \text d\tilde U \\
		& = f_u(U; x, t).
\end{split}
\end{align}  
In words, the ensemble average of $\pi_u$ coincides with the single-point PDF of $u(x,t)$. This suggests a two-step procedure for derivation of PDF equations. First, one derives an equation for $\pi_u(U-u)$. Second, one ensemble-averages (homogenizes) the resulting equation to obtain a PDF equation.

The first step relies on rules of differential calculus applied, in the sense of distributions, to the function $\pi_u(U-u)$,
\begin{equation}
  \frac{\partial \pi_u}{\partial u} = - \frac{\partial \pi_u}{\partial U}, \qquad
  \frac{\partial \pi_u}{\partial t} = \frac{\partial \pi_u}{\partial u} \frac{\partial u}{\partial t} = - \frac{\partial \pi_u}{\partial U} \frac{\partial u}{\partial t}, \qquad
  \frac{\partial \pi_u}{\partial x} = - \frac{\partial \pi_u}{\partial U} \frac{\partial u}{\partial x}.
\end{equation}
Multiplying both sides of~\eqref{linadv} with $\partial_U \pi_u$, using the above relations and the sifting property of the delta function, $g(u) \delta(U-u) = g(U) \delta(U-u)$ for any ``good'' function $g(u)$, we obtain a linear stochastic PDE for $\pi_u$,
\begin{align}\label{eq:pi}
\frac{\partial \pi_u}{\partial t} + k \frac{\partial \pi_u}{\partial x} + \frac{\partial g(U) \pi_u}{\partial U} = 0.
\end{align}

The second step, ensemble averaging of this equation, is facilitated by Reynolds decomposition that represents all the independent and dependent variables involved as the sums of their ensemble means and zero-mean fluctuations about these means, i.e., $k = \bar k + k'$ and $\pi_u = f_u + \pi_u'$ with $\mathbb E[k'] = 0$ and $\mathbb E[u'(x,t)] = 0$. Then, the ensemble average of~\eqref{eq:pi} yields an unclosed PDE for $f_u(U;x,t)$,
\begin{align}
%- \frac{\partial \pi_u}{\partial t} &- (\bar k + k') \frac{\partial \pi_u}{\partial x} = 0 \\
%\frac{\partial \pi_u}{\partial t} & + \bar k \frac{\partial \pi_u}{\partial x} + k' \frac{\partial \pi_u}{\partial x} = 0 \\
\frac{\partial f_u}{\partial t} + \bar k \frac{\partial f_u}{\partial x} + \frac{\partial g(U) \pi_u}{\partial U} + \mathcal M(f_u) = 0, \qquad \mathcal M(f_u) \equiv \mathbb E\left[ k' \frac{\partial \pi_u'}{\partial x} \right] = \frac{\partial \mathbb E[k'\pi_u']}{\partial x};
\end{align}
which is the same as~\eqref{margpdfeq}. %a conservation law of the form
%\begin{equation}
%\frac{\partial f_u}{\partial t} + \frac{\partial }{\partial x} ( \bar k f_u + \langle k' \pi_u' \rangle) = 0 
%\end{equation}
%where $\langle k' \pi_u' \rangle$ is a closure term to be approximated.

%%%%%%%%%%%%%%%%%%%%%%%%%%%%%%%%%%%%%%%%%%%%%%%%%%%
\section{Derivation of the Joint PDF equation}
\label{jpdfderivation}
%%%%%%%%%%%%%%%%%%%%%%%%%%%%%%%%%%%%%%%%%%%%%%%%%%%

%Similarly, it can be shown that for two random fields $u(x, t)$ and $v(x, t)$ we obtain a joint PDF by average the product of raw PDFs, s.t. $\langle \pi_u \pi_v \rangle = f_{uv}(U, V; x, t)$.

Consider a generalized function
\begin{align}
\pi_{uk}(U-u,K-k) = \delta(U - u(x,t)) \delta(K - k).    
\end{align} 
Let $f_{uk}(U, K;x,t)$ denote a joint PDF of the random input $k$ and the random output $u$ at any space-time point $(x,t)$. Then, in analogy to~\eqref{eq:avpi}, $\mathbb E[\pi_{uk}] = f_{uk}(U, K; x, t)$. A procedure similar to that used to derive a stochastic PDE~\eqref{eq:pi} now yields a deterministic PDE for $\pi_{uk}$,
\begin{align}%\label{eq:pi}
\frac{\partial \pi_{uk}}{\partial t} + K \frac{\partial \pi_{uk}}{\partial x} + \frac{\partial g(U) \pi_{uk}}{\partial U} = 0.
\end{align}
The randomness of $\pi_{uk}$ stems from the random initial state $u_0$, rather than the model coefficients. Consequently, the averaging of this equation is trivial and exact, and given by~\eqref{eq:fuk}.
%\begin{align}%\label{eq:pi}
%\frac{\partial f_{uk}}{\partial t} + K \frac{\partial f_{uk}}{\partial x} + \frac{\partial g(U) f_{uk}}{\partial U} = 0.
%\end{align}
This equation is subject to the initial condition $f_{uk}(U, K; x, 0) = f_{u_0,k}(U, K; x)$. If $u_0(x)$ and $k$ are mutually independent, then $f_{uk}(U, K; x, 0) = f_{u_0}(U; x) f_k(K)$.
%\begin{equation}
%	\frac{\partial \prod_{i} \pi_i}{\partial x} = \sum_j \prod_{i \neq j} \pi_i \frac{\partial \pi_j}{\partial x}, 
%\end{equation}

\textbf{[Fix the rest]} The solution to this equation can be obtained analytically using the method of characteristics
\begin{equation}
f_{uk}(U, K; x, t) = f_{u_0}(U; x - Kt) f_k(K).
\end{equation}
%The corresponding marginal distribution in $u$ only is given by the integral $f_u(U; x, t) = \int_{-\infty}^{+\infty} f_{uk}(U, K; x, t) dK$.

%%%%%%%%%%%%%%%%%%%%%%%%%%%%%%%%%%%%%%%%%%%%%%%%%%%
\section{Derivation of Closure Approximations}
\label{app:nonlocal}
%%%%%%%%%%%%%%%%%%%%%%%%%%%%%%%%%%%%%%%%%%%%%%%%%%%

One way to solve for the equation for higher moments is to perturb $\pi_u = f_u + \pi_u'$ before averaging, giving the equation
\begin{equation}\label{perteq}
\frac{\partial \pi_u'}{\partial t} + \frac{\partial f_u}{\partial t} + k' \frac{\partial \pi_u'}{\partial x} + k' \frac{\partial f_u}{\partial x} + \bar k \frac{\partial \pi_u'}{\partial x} + \bar k \frac{\partial f_u}{\partial x}
\end{equation}
Subtracting the marginal PDF Eqn.~\ref{margpdfeq} from Eqn.~\ref{perteq}, multiplying by $k'$ and ensemble averaging, we get
\begin{equation*}
\frac{\partial \langle k' \pi_u' \rangle}{\partial t} + \frac{\partial \langle k' k' \pi_u' \rangle}{\partial x} + \langle k' k' \rangle \frac{\partial f_u}{\partial x} + \bar k \frac{\partial \langle k' \pi_u' \rangle}{\partial x} - \frac{\partial \langle k' k' \pi_u' \rangle}{\partial x} = 0
\end{equation*}

Third order terms are assumed to be small, and $\sigma_k^2 \equiv \langle k' k' \rangle$. Finally, we get
\begin{equation}\label{closureeqn}
\frac{\partial \langle k' \pi_u' \rangle}{\partial t} + \bar k \frac{\partial \langle k' \pi_u' \rangle }{\partial x} + \sigma_k^2 \frac{\partial f_u}{\partial x} = 0
\end{equation}
which is an advection PDE for the closure term $\langle k' \pi_u' \rangle$. We now have two equations (Eqn.~\ref{margpdfeq} and \ref{closureeqn}), and two unknowns: $f_u$ and $\langle k' \pi_u' \rangle$.

If the third term in Eq.~\ref{closureeqn} is taken as a forcing term, the corresponding Green's function solution of the equation is
\begin{equation}
\langle k' \pi_u' \rangle  = - \sigma_k^2 \int_0^t \int_{-\infty}^{+\infty} G(x, y, t, \tau) \frac{\partial f_u(U; y, \tau)}{\partial y} dy d\tau 
\end{equation}\label{closureint}
where the Green's function is $G(x, y, t, \tau) = \delta(x - y - \bar k(t - \tau))$, thus
\begin{align*}
\langle k' \pi_u' \rangle(U; x, t)  &= - \sigma_k^2 \int_0^t \int_{-\infty}^{+\infty} \delta(x - y - \bar k(t - \tau)) \frac{\partial f_u(U; y, \tau)}{\partial y} dy d\tau \\
&= - \sigma_k^2 \int_0^t \frac{\partial f_u(U; y = x - \bar k(t - \tau), \tau)}{\partial y} d\tau.
\end{align*}
This yields~\eqref{eq:nonlocal}.

We localize in time, resulting in the following expression
\begin{equation}
\langle k' \pi_u' \rangle(U; x, t) \approx - \hat \sigma_k^2 \frac{\partial}{\partial x} \int_0^t f_u(U; x, t) d\tau 
\end{equation}
which, when combined with Eq.~\ref{margpdfeq} and differentiated in time, becomes
\begin{equation}\label{2ordereq}
\frac{\partial^2 f_u}{\partial t^2} + \hat{\bar{k}} \frac{\partial^2 f_u}{\partial x \partial t} - \hat \sigma_k^2 \frac{\partial ^2 f_u}{\partial x^2} = 0
\end{equation}
Note the presence of a second order time derivative and a mixed derivative. 
This equation is a wave equation with two wave speeds $v_{\pm} =  \tfrac{1}{2} \left( \hat{\bar{k}} \pm  \sqrt{ \hat{\bar{k}}^2 + 4 \hat{\sigma}_k^2 }\right)$, which can be factored as a system of the form
\begin{align*}\label{waveform}
\frac{\partial f_u}{\partial t} &+ v_+ \frac{\partial f_u}{\partial x} = f'_u \\
\frac{\partial f'_u}{\partial t} &+ v_- \frac{\partial f'_u}{\partial x} = 0
\end{align*}

\bibliographystyle{abbrvnat}
\bibliography{%mlpde,
mlpde2}

\end{document}